\documentclass[sigconf]{acmart}
\usepackage{xcolor}
\usepackage{mwe}
\usepackage{caption}
\usepackage{subcaption}
\usepackage{amsmath}
\usepackage{algorithmicx}
\usepackage{algorithm}
\usepackage{algpseudocode}

\setcopyright{none}
\renewcommand\footnotetextcopyrightpermission[1]{} 
\AtBeginDocument{%
  \providecommand\BibTeX{{%
    \normalfont B\kern-0.5em{\scshape i\kern-0.25em b}\kern-0.8em\TeX}}}

\acmConference[ArXiv]{ArXiv}{August 14-18, 2022}{Washington, D.C.}

\begin{document}
\pagestyle{empty}

\title{Multi-objective Optimization of Notifications Using Offline Reinforcement Learning}

\author{Prakruthi Prabhakar}
\affiliation{%
  \institution{LinkedIn Corporation}
  \city{Mountain View, CA}
  \country{USA}
}
\email{paprabhakar@linkedin.com}

\author{Yiping Yuan}
\affiliation{%
  \institution{LinkedIn Corporation}
  \city{Mountain View, CA}
  \country{USA}
}
\email{ypyuan@linkedin.com}

\author{Guangyu Yang}
\affiliation{%
  \institution{LinkedIn Corporation}
  \city{Mountain View, CA}
  \country{USA}
}
\email{guyang@linkedin.com}

\author{Wensheng Sun}
\affiliation{%
  \institution{LinkedIn Corporation}
  \city{Mountain View, CA}
  \country{USA}
}
\email{wesun@linkedin.com}

\author{Ajith Muralidharan}
\affiliation{%
  \institution{LinkedIn Corporation}
  \city{Mountain View, CA}
  \country{USA}
}
\email{amuralidharan@linkedin.com}
\renewcommand{\shortauthors}{Prakruthi Prabhakar et al.}

\begin{abstract}
Mobile notification systems play a major role in a variety of applications to communicate, send alerts and reminders to the users to inform them about news, events or messages. In this paper, we formulate the near-real-time notification decision problem as a Markov Decision Process where we optimize for multiple objectives in the rewards. We propose an end-to-end offline reinforcement learning framework to optimize sequential notification decisions. We address the challenge of offline learning using a Double Deep Q-network method based on Conservative Q-learning that mitigates the distributional shift problem and Q-value overestimation. We illustrate our fully-deployed system and demonstrate the performance and benefits of the proposed approach through both offline and online experiments.


\end{abstract}



\maketitle

\section{Introduction}
Notifications play an important role for mobile applications to keep users informed and engaged. With the right content at the right time, notifications can inform users of important activities and bring more value to users. Through this, notifications also help increase user engagements with the platform. 

Mobile applications mainly serve two categories of notifications to users. The first category is near-real-time notifications \cite{gao2018near}, which are sent close to the event generation time, and are usually triggered by activities on the platform. They are time-sensitive and need to be delivered in near-real-time due to their nature, user expectation, or product constraints. Examples are notifications about a live video stream, or a recent share on the platform or a new job post that the user wanted to be informed about. The second category is offline (batch) notifications \cite{yuan2019state}, which are generated offline usually in batches. They are not time-sensitive and would be relevant if delivered within a predefined time window. Examples of these notifications include events about your network, such as your colleague's work anniversary and birthday, or aggregate notifications about activities that you may be interested in. In this paper, we focus our discussion on a near-real-time notification system, which can process both near-real-time and offline notifications and make decisions in a stream fashion in near-real-time. An example of such a distributed near-real-time notification system can be found in \cite{gao2018near}.  Note that a near-real-time notification system can process offline notifications and spread them out over time, for example using a notification spacing queuing system introduced in \cite{yuan2019state}. On the other hand, a system designed solely for offline notifications may not be able to process near-real-time notifications. 

There are a few characteristics of the notification system that make them suitable applications for reinforcement learning (RL). Sending a notification may lead to two categories of feedback:  notification content engagement responses (e.g., clicks, dismisses) and site engagement responses (e.g., user visits, notification disables). First, these responses are usually delayed rewards in the sense that a user may visit the site a few hours or days after receiving a notification. Second, user engagement responses, particularly the site engagement responses, usually cannot be attributed to a single notification, but rather a sequence of notification decisions, presenting an attribution challenge for supervised modeling approaches. Third, optimal notification user experience depends on a sequence of well-coordinated notification decisions in terms of notification spacing and frequency. Last but not the least, a notification system not only cares about the immediate engagement, i.e., the next user visit, but also the long-term user engagement \cite{zou2019reinforcement}, which can be measured by the aggregated rewards over time. A greedy approach to optimize the notification decisions for the immediate engagement can harm the long-term engagement.

RL has been an active research area for decades, that provides a mathematical framework for learning-based control. By utilizing RL and combining with deep neural network function approximators, we can automatically acquire near-optimal behavioral skills, represented by policies, for optimizing user-specific reward functions, and get generalizable and powerful decision-making engines. RL algorithms can operate well in scenarios where we do not have well understood models of the phenomenon it seeks to optimize. For large-scale online systems, online RL training with online exploration may not be feasible due to high infrastructure costs, unknown time to convergence, and unbounded risks of deteriorating user experience.
Inspired by the success of machine learning methods which train on large-scale labeled datasets, data-driven reinforcement learning or offline reinforcement learning (offline RL) has become an active research area ~\cite{lange2012batch, chen2019top,ie2019reinforcement,fujimoto2019off,levine2020offline, kumar2020conservative}.  There are theoretical and practical challenges in efficient offline policy learning due to the notorious Deadly Triad problem (i.e., the problem of instability and divergence
arising when combining function approximation, bootstrapping and offline training) \cite{sutton1998reinforcement,van2018deep}. Accurate offline policy evaluation which is an important component of offline learning is also challenging\cite{levine2020offline, mahmood2014weighted,thomas2016data,xie2019towards}. Early works started with finding suitable existing online algorithms to be  converted to offline. Recently, algorithms such as conservative Q-learning (CQL)~\cite{kumar2020conservative} and implicit Q-learning~\cite{kostrikov2021offline} are proposed to address unique offline learning challenges. In this paper, we explore both Offline Double Deep Q-networks (Offline DDQN) in the first category and its CQL version in the second category. We demonstrate how CQL improves offline learning through offline and online experiments. Offline evaluation is another key component to the offline RL framework. We rely on a one-step importance sampling method to evaluate trained policies for offline experiments and for choosing online test candidates.

One practical challenge for applying RL is that real-world applications usually optimize for multiple objectives rather than a single scalar reward. At LinkedIn, notifications drive a large number of user and business objectives \cite{gao2018near, amuralidharan2022}. Engagement on the notification card helps drive conversations and provide measurable objectives that capture user satisfaction. On the other hand, notifications provide top of the funnel growth opportunities for user visits and further interactions onsite. In other scenarios, businesses may want to separately optimize for engagements on a subset of notifications (e.g. engagement on jobs related notifications). Finally, we often have guardrails on notification volume or disables. A simple decision system, which optimizes for one objective, will fail to account for these multiple objectives without the use of extensive heuristics. Additionally, it will fail to find opportunities to provide the best trade-offs between core metrics and guardrails. 

In this paper, we apply a multi-objective reinforcement learning (MORL) framework through reward engineering~\cite{zou2019reinforcement,silver2021reward}. We compare this with the state-of-the-art multi-objective optimization (MOO) on top of supervised learning models described in \cite{amuralidharan2022, gao2018near} and show that such an RL framework performs better in the online A/B test. This RL framework is fully deployed in LinkedIn. 

Finally, we suggest a potential opportunity to combine the power of RL and supervised learning by leveraging reward prediction models in offline training. This can be seen as a hybrid model-based reinforcement learning~\cite{kaiser2019model}. We show that using predicted rewards instead of the observed rewards may improve the offline learning if it gives better bias-variance trade-offs. Since most existing notification systems are built on supervised prediction models, such a connection could smooth the transition into an RL framework for industry applications. 




\section{Related Work}
\label{sec:relatedWork}

Most notification systems \cite{gupta2017optimizing,yuan2019state,gao2018near,zhao2018notification} are built around response prediction models. These response predictions are then compared with optimal thresholds from online or offline threshold search based on  multi-objective optimization \cite{agarwal2011click}. While such systems have demonstrated good empirical performance, they could  be sub-optimal in decision-making. A bandit-based solution \cite{wu2017returning} was proposed to improve long-term user engagement in a recommender system. Recent work in \cite{yuan2022notification} applied RL to a special notification spacing system for offline notifications, which does not extend to near-real time notifications with variable decisions times. 

Offline RL has been widely applied in domains such as robotics~\cite{dasari2019robonet}, autonomous driving~\cite{kiran2021deep},  advertising and recommender systems~\cite{chen2019top, ie2019reinforcement}, and language and dialogue applications ~\cite{jaques2019way}. Chen et al. \cite{chen2019top} applied a Policy Gradient learning in YouTube recommender system with Off-policy correction for offline RL. Ie et al. \cite{ie2019reinforcement} proposed an offline RL approach for slate-based recommender systems, where the large action space can become intractable for many RL algorithms. Zou et al. \cite{zou2019reinforcement} designed an offline framework to learn a Q-network and a separate S-network from a simulated environment to assist the Q-network.  While there are a lot of efforts to apply RL to real-world applications \cite{zhao2018recommendations, wang2020incremental}, driving long-term engagement through notification systems 
 presents its unique challenges and opportunities for RL due to its sequential planning and short-term long-term trade-offs.

Despite the advantages of existing offline RL approaches, in practice, offline RL presents a major challenge. Standard off-policy RL methods could fail due to overestimation of values induced by the distribution shift between the learned policy and the dataset \cite{levine2020offline}. To tackle the limitations, Conservative Q-learning (CQL) was proposed~\cite{kumar2020conservative}, and was applied to many different domains, such as voice assistants~\cite{bayramouglu2021engagement} and robotics~\cite{kumar2020conservative}.

\section{Near-real-time Notification System}

A general near-real-time notification system takes a stream of notification requests and makes near-real-time decisions on whether the notification should be sent or not. A system at scale may process hundreds of notification requests for each of millions of users. As we mentioned before, such a system is usually optimized for both notification content engagement responses and site engagement responses.

\subsection{Markov Decision Process}

The key concepts for the notification decision problem are described below.

\textbf{Actions.} $a$ denotes an action in the action space $\mathcal{A}$. We consider a discrete action space consisting of two actions - SEND (send the notification candidate to the user) and  NOT-SEND. For near real time notifications, NOT-SEND ends up dropping the notification, while for offline notifications we queue the notification for another evaluation later. 

\textbf{States.} $s$ denotes a state in the state space $\mathcal{S}$. A state represents a situation in the environment and summarizes all useful historical information. States must be defined properly to ensure the Markov property. In our problem setting, we use a plethora of features, including the number of notifications sent to the user since the last visit, number of notifications sent in the past day, user profile information and contextual user information to represent a state. In this manner, we allow the user to be part of the environment and represent their interests and context using a rich state representation.

\textbf{Environment.} In standard RL, an agent interacts with an environment over a number of discrete time steps. At every time step $t_k$, the agent receives a state $s_k$ and chooses an action $a_k$. In return, the agent receives the next state $s_{k+1}$ and a scalar reward $r_k$. In our problem, the environment captures the distribution of all users' interests and interactions. A single episode corresponds to a sampled user and a sequence of candidate notifications evaluated to be sent or not-sent to the user. 

\textbf{Reward.} $r_k$ denotes the reward observed after taking action $a_k$. In our notifications system, we measure multiple rewards - interactions on the notification cards (clicks); user visits (sessions); volume penalty (represented as negative of volume of notifications as a guardrail). Notification clicks and volume penalty are directly attributed to the notification send decision, and these are 0 when notifications are not sent. For sessions reward, we measure the number of user visits between two notification decision times $t_k$ and $t_{k+1}$. With notifications application, we cannot choose to optimize a single positive reward as that trivially results in poor performance on the guardrails. We discuss in detail how to choose a linear combination of rewards in Section~\ref{sec:morl}. Using the final scalar reward, the total return $R_k = \sum_{s=k}^\infty \gamma^k r_k$ represents the time-discounted total reward for a user. Here, $\gamma \in (0,1]$ is the discount factor, which controls the trade-offs between the short-term and long-term rewards. The goal of the agent is to maximize this total return to encourage long-term user engagement. 

\textbf{Policy.} A policy $\pi$ is a mapping from the state space to the action space. In this setting, it makes SEND or NOT-SEND decisions given the state features. A policy can be either deterministic or stochastic. For every MDP, there exists an optimal deterministic policy $\pi^*$, which maximizes the total return from any initial state.

The final goal in an RL problem is to learn a policy, which explicitly or implicitly defines a distribution $\pi({a}|s)$ over actions conditioned on states. Using the Markovian assumption, we can derive the trajectory distribution of $p_{\pi}$, where the trajectory is a sequence of states and actions of length H, given by $\tau=(s_0, {a}_0, \dots, s_H, {a}_H)$

\begin{equation}
p_{\pi}(\tau)=d_{0}\left({s}_{0}\right) \prod_{k=0}^{H} \pi\left({a}_{k} \mid {s}_{k}\right) T\left({s}_{k+1} \mid {s}_{k}, {a}_{k}\right),
\end{equation}

where $d_{0}\left({s}_{0}\right)$ is the initial state distribution,  $T\left({s}_{k+1} \mid {s}_{k}, {a}_{k}\right)$ is the transition probability to ${s}_{k+1}$ given ${s}_{k}, {a}_{k}$ . The RL objective, $J(\pi)$, can then be written as,

\begin{equation}\label{eqn:job_J}
J(\pi)=\mathbb{E}_{\tau \sim p_{\pi}(\tau)}\left[\sum_{k=0}^{H} \gamma^{k} r_k\right].
\end{equation}

\subsection{Time Discretization}
\label{sec:discretization}
\begin{figure}
\begin{subfigure}{.5\textwidth}
  \centering
  \includegraphics[width=.77\linewidth]{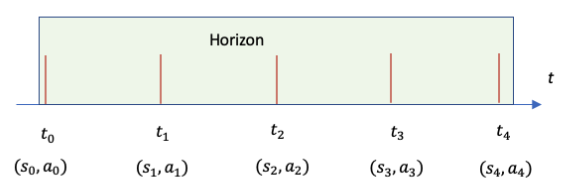}  
  \caption{Uniform time-step}
  \label{fig:sub-fixed}
\end{subfigure}

\begin{subfigure}{.5\textwidth}
  \centering
  \includegraphics[width=.8\linewidth]{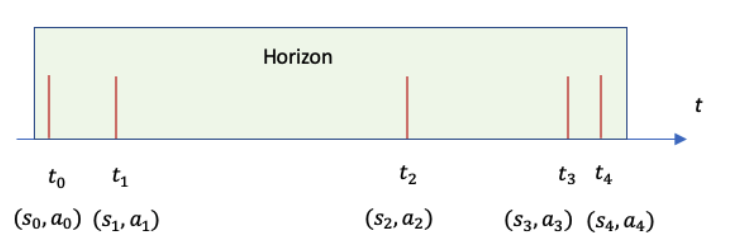} 
  \caption{Non-uniform time-step}
  \label{fig:sub-variable}
\end{subfigure}
\caption{Discrete time-steps}
\label{fig:fig}
\end{figure}
RL problems are often formulated into either a discrete time sequence, where at each time step, states, rewards are observed and actions are made; or a continuous control system, where actions, states and rewards are all functions of a continuous time. It is straightforward to convert a continuous state system into a uniform discrete state system when the evaluation time steps are periodic as in Figure \ref{fig:sub-fixed}.

In real-world applications though, the environment may be somewhat in between. Many RL applications on recommendation systems correspond to an agent evolving in the continuous environment, but only making observations and choosing actions at $t_k$ with irregular intervals, as illustrated in Figure \ref{fig:sub-variable}. We refer to this as a non-uniform discrete environment as oppose to the typical uniform discrete environment in Figure \ref{fig:sub-fixed}. In the near-real-time notification system,  we have a stream of incoming notification candidates for each user being evaluated at previously unknown arbitrary timestamps. We only make decisions at the arrival timestamp of these candidate notifications but reward and state changes may happen at any time. 

A non-uniform time-step discretization allows the time span of each step to be different. Rewards collected between $t_k$ and $t_{k+1}$ are attributed to reward $r_k$. The long-term reward can be approximately represented as $R(\tau) = \sum_{k=0}^{H} \gamma^{t_{k+1}-t_0} r_k$ with a discount factor applied over time. This approximation happens, because we do not distinguish between different times the reward occurs within two discrete time steps. This is typically a reasonable assumption if the reward occurrence is equally distributed within any time period.

\section{Methodology}
\label{sec:methodology}

\subsection{Offline Reinforcement Learning}
\label{sec:offlineRL}
In general, there are two kinds of RL approaches. The first approach is the online on-policy reinforcement learning, where the policy $\pi$ is learned from data collected using the policy $\pi$ itself. The other approach is off-policy reinforcement learning, where the agent's experience can be collected using any policy with sufficient exploration. These two RL paradigms involve iteratively collecting data by interacting with the environment through the latest learned policy and using experience to improve the policy. However, this process is sometimes expensive, risky, and potentially dangerous, especially if the domain is complex and effective generalization requires a large amount of data. For instance, a healthcare agent interacts with the environment using a partially trained policy, which might take potentially unsafe actions, administering a wrong drug to a patient. A partially trained RL agent in large-scale notification system could send inappropriate or excessive notifications to users, which would damage trust and long-term user engagement. Unlike on-policy RL and off-policy RL that are difficult to reconcile with the data-driven paradigm, offline reinforcement learning (offline RL) can be defined as a data-driven formulation of the RL problem, utilizing only previously collected offline data, without any additional online interaction~\cite{levine2020offline}.  Offline RL is a much safer choice, because the policy is only deployed after being fully trained. Additionally, it can reuse previously collected data from existing policies. 

\subsection{Double Q-Learning}
\label{sec:dqn}
Before diving into offline RL, we start from an important off-policy RL approach, the Q-learning and its variant Double Q-learning. Q-learning is one of the most popular RL algorithms~\cite{sutton1998reinforcement} for estimating optimal action values. A deep Q network (DQN) is a multi-layered neural network with the state $s$ as inputs and outputs a vector of action values $Q(s, *; \theta)$~\cite{van2016deep}. The target of DQN is

\begin{equation}
Y_k^{DQN} = r_{k} + \gamma ^{t_{k+1}-t_{k}} \underset{\mathbf{a}}{\mathrm{max}}\ Q(s_{k+1}, \mathbf{a}; \theta_{k}),
\end{equation}

which can be refactored as,

\begin{equation}\label{eqn:modified_dqn}
Y_k^{DQN}=r_{k} + \gamma^{t_{k+1}-t_{k}} Q(S_{k+1}, \underset{\mathbf{a}}{\mathrm{argmax}}\ Q(s_{k+1}, \mathbf{a}; \theta_{k}); \theta_k).
\end{equation}

When taking action $a_k$ in state $s_k$ and observing the immediate reward $r_{k}$ and resulting state $s_{k+1}$, the Bellman error over the mini-batch, $\mathcal{E}(B, \theta_k) = \sum(Q_k - Y_k^{DQN})^2$ can be calculated and the parameters $\theta_ k$ can be updated by the mini-batch stochastic gradient descent. The Q-learning algorithm tends to be over-optimistic and learns unrealistically high action values in large-scale problems, because the maximization step tends to prefer overestimated action values to underestimated ones~\cite{van2016deep}. To reduce over estimation, Mnih et al.~\cite{mnih2015human} proposed a Double DQN (DDQN) that introduced a target network which decomposes the $\mathrm{max}$ operation in the target into action selection and action evaluation using separate networks. The target network is the same as DQN except that the network weights $\theta^{-}$ are copied every $n$ steps from the DQN, $\theta^{-}=\theta$. 

Replacing the outer Q network by the target network $Q'$ in \eqref{eqn:modified_dqn}, we have the new target,

\begin{equation}\label{eqn:ddqn}
Y_k^{DDQN}=r_{k} + \gamma^{t_{k+1}-t_{k}} Q'(s_{k+1}, \underset{\mathbf{a}}{\mathrm{argmax}}\ Q(s_{k+1}, \mathbf{a}; \theta_k);\theta^{-}_k).
\end{equation}

Here, we estimate the action corresponding to the current parameters of DQN ($Q(.,.;\theta_k)$), and evaluate the target using the target network ($Q(.,.;\theta_k')$).

\subsection{Conservative Q-Learning}
Offline RL uses previously collected data from existing policies to obtain policies that optimize the objective $J(\pi)$ (Eq. \ref{eqn:job_J}). However, it's a more difficult problem for multiple reasons. First, the learning algorithm must rely entirely on the static dataset of transitions $\mathcal{D}=\left\{\left(s_{k}^{i}, a_{k}^{i}, s_{k+1}^{i}, r_{k}^{i}\right)\right\}$ and there is no way to interact with the environment and collect additional transitions using the behavior policy. Therefore, we are unable to discover the high-reward regions that are not covered by dataset $\mathcal{D}$. Additionally, the offline dataset is collected from a distribution which may be very different from the one encountered when deploying the learned model online. Major RL algorithms, e.g. the Double Q-learning, can not learn effectively from entirely offline data due to such a distribution shift.

Directly utilizing existing value-based off-policy RL algorithms in an offline setting results in poor performance due to issues with bootstrapping from out-of-distribution actions and over-fitting. To tackle the distribution shift presented in offline RL, a number of offline methods are being developed~\cite{fu2019diagnosing,siegel2020keep, levine2020offline}. In general, we regularize either the value function~\cite{kostrikov2021offline} or Q-function~\cite{kumar2020conservative} to avoid overestimation for out-of-distribution actions. In this paper, we adopt and implement the conservative Q-learning (CQL) for learning conservative, lower-bound estimates of the value function, by regularizing the Q-values during training. Kumar et al~\cite{kumar2020conservative} ensures a conservative Q-function by modifying the objective for fitting the Q-function. Specifically, it adds an additional conservative penalty term to the objective,

\begin{equation}\label{eqn:cql_error}
\mathcal{E}^{CQL}(B, \theta)=\alpha \mathcal{C}(B, \theta)+ \frac{1}{2} E_{s,a,s' \sim D}(Q(;\theta) - Y^{DDQN}_k )^2
\end{equation}

where $\mathcal{C}(B, \theta)$ is the CQL penalty term, $B$ is the behavior policy and $\alpha$ is the control parameter. Different choices for $\mathcal{C}(B, \theta)$ lead to algorithms with different properties and the following $\mathcal{C}(B, \theta)$ is for Q-learning,

\begin{multline}\label{eqn:cql_error_2}
\mathcal{C}(B, \theta)= \max_{\mu} ( R(\mu) + \mathbb{E}_{\mathbf{s} \sim B, \mathbf{a} \sim \mu(\mathbf{a} \mid \mathbf{s})}\left[Q(\mathbf{s}, \mathbf{a}; \theta)\right] \\
-\mathbb{E}_{(\mathbf{s}, \mathbf{a}) \sim B}\left[Q(\mathbf{s}, \mathbf{a}; \theta)\right]).
\end{multline}

Here, $\mu$ is the policy we are searching for, given the network structure described by equation \eqref{eqn:ddqn}. When we choose the regularizer function $R(\mu)$ as $R(\mu) = -D_{KL}(\theta|unif)$, where $D_{KL}$ is the Kullback–Leibler divergence and  $unif$ is the uniform prior distribution, we get the following term $\mathcal{C}(B, \theta)$

\begin{equation*}
\mathcal{C}(B, \theta) =  \mathbb{E}_{\mathbf{s} \sim \mathcal{B}}\left[\log \sum_{\mathbf{a}} \exp (Q(\mathbf{s}, \mathbf{a}))-\mathbb{E}_{\mathbf{a} \sim \hat{\pi}_{\beta}(\mathbf{a} \mid \mathbf{s})}[Q(\mathbf{s}, \mathbf{a})]\right].
\end{equation*}

This conservative penalty minimizes Q-values under the adversarially chosen $\mu(\mathbf{a}|\mathbf{s})$ distribution, and maximizes the values for state-action tuples in the batch, which ensures the high Q-values are only assigned to in-distribution actions.

\subsection{Practical Algorithm}
We now describe our offline double Q-learning method based on CQL. Pseudocode is shown in Algorithm~\ref{alg:DQL}. 

\begin{algorithm}
\caption{Double Q-learning with CQL}\label{alg:DQL}
\begin{algorithmic}[1]
\item Initialize DQN network Q with random weights $\theta$
\item Initialize target network $Q^{-}$ with weights $\theta^{-} = \theta$
\item N <- (size of dataset $D$ / batch size $m$) $\times$ number of epochs $E$
\For{\texttt{step=1:N}}
    \State \texttt{Get a random batch of size $m$ from offline dataset $D$}
    \State \texttt{Compute empirical CQL error $\mathcal{E}^{CQL}(B, \theta)$ through equation \eqref{eqn:cql_error}}
    \State \texttt{Perform a minibatch stochastic gradient descent step with respect to $\theta$} to minimize the CQL error
    \State \texttt{Every $n$ steps update target network with $\theta^{-} = \theta$}
\EndFor
\end{algorithmic}
\end{algorithm}


\subsection{Multi-objective Reinforcement Learning}
\label{sec:morl}
In real-world applications, it is usually the case that we need to optimize the system for multiple objectives. For the notification system, we need to balance content engagement and site engagement objectives while avoiding sending too many notifications. Multi-objective reinforcement learning (MORL) deals with learning control policies to simultaneously optimize over several criteria. Compared to traditional RL aiming to optimize a scalar reward, the optimal policy in a multi-objective setting depends on the relative preferences among potentially competing criteria. The MORL framework provides many advantages over traditional RL, but learning policies over multiple preferences under the MORL setting is quite challenging.  We utilize the following approach:  (1) convert the multi-objective problem into a single scalar reward through a weighted linear combination (i.e., a preference function) similar to how its represented in \cite{mossalam2016multi}; and (2) train a set of optimal policies in parallel that encompass the space of possible preferences and pick the best policy through offline evaluation. Linear preferences are intuitive and suitable for our use case as we can to value different metrics (and the trade-offs) on a common scale, with the weights representing the value for each unit change in the metric. The different objectives could be contradictory in nature to one another, and the right chosen preferences helps the network learn to balance these objectives and their trade-offs. An initial set of preferences can be obtained from a combination of previous experiments analysis, as well as causal and correlational analyses mapping metric impact towards long-term true north goals. The reward at time $t_k$ is
\begin{equation}
\label{eq:linear_rewards}
    r_k = \boldsymbol{\omega}^T \boldsymbol{m}_k,
\end{equation}
where $\boldsymbol{\omega}$ is the preference vector consisting of weights on each objective, and $\boldsymbol{m}$ is the reward vector. For this notification application, we represent $\boldsymbol{m}_k$  by the following objectives,
\begin{itemize}
    \item $m^{s}_k$ is the number of site visits between time $t_k$ and time $t_{k+1}$. This is an example of site engagement responses.
    \item $m^{c}_k$ is $1$ if there is a click on the notification sent at $t_k$; $0$ if there is no click on the notification sent at $t_k$ or if the notification is not sent at $t_k$. This is an example of notification content engagement responses.
    \item $m^{v}_k$ is $-1$ if the action is SEND and $0$ if the action is NOT-SEND. This negative reward serves as a notification volume penalty, since we want to minimize the volume of the notifications.
\end{itemize}
So the reward at time $t_k$ is
\begin{equation}
\label{eq:linear_rewards_breakdown}
    r_k = \omega^{s}_k m^{s}_k + \omega^{c}_k m^{c}_k + \omega^{v}_k m^{v}_k.
\end{equation}
Another characteristic that separates $m^{s}_k$ and $m^{c}_k$ from $m^{v}_k$ is that $m^{s}_k$ and $m^{c}_k$ are stochastic rewards, meaning that their values are probabilistic conditional on the action $a_k$ and state $s_k$. Such stochastic rewards may be noisy and it is especially the case for $m^{s}_k$. A user's site engagement depends on a lot of other factors other than their notification experience. A user may have site visits following a NOT-SEND action and no site visits following a SEND action. For stochastic rewards, we may consider using the predicted rewards from  a supervised model $\hat{E}(m_k|a_k,s_k)$ instead of the observed rewards in the offline training. While a prediction model $\hat{E}(m_k|a_k,s_k)$ will inevitably bring a modeling bias, it can reduce the variance of the offline learning. A better bias-variance trade-off could potentially lead to a more efficient policy learning. For existing notification systems, these reward prediction models are usually already available. In practice, the decision to use a predicted reward depends on the stochastic nature of the reward, the accuracy of the reward prediction model, as well as the sparsity level of the reward. In this paper, we use $\hat{E}(m^{c}_k|a_k,s_k)$ from an XGboost model ,which is already available in the system, instead of $m^{c}_k$ in the reward formulation,
\begin{equation}
\label{eq:linear_rewards_prediction}
    r_k = \omega^{s}_k m^{s}_k + \omega^{c}\hat{E}(m^{c}_k|a_k,s_k)  + \omega^{v}_k m^{v}_k.
\end{equation}
This is because notification clicks are clean to attribute and their prediction models are fairly mature. Additionally, in section~\ref{sec:online_predicted_delta_pvisit}, we provide further experiments with predicted $\hat{E}(m^{s}_k|a_k,s_k)$ instead of observed $m^{s}_k$ . 

\subsection{Off-policy Evaluation via Importance Sampling}
\label{sec:offlineEvaluation}

To evaluate the new policy trained from the offline dataset $D$, off-policy evaluation (OPE) is used~\cite{precup2000eligibility}. OPE provides typically unbiased or strongly consistent estimators~\cite{precup2000eligibility}. For instance, we can use importance sampling to derive an unbiased estimator of $J(\pi_{\theta})$ by correcting the mismatch in the distribution under the behavior policy and the target policy. Then \eqref{eqn:job_J} can be modified as

\begin{align}
 \begin{aligned}
J\left(\pi_{\theta}\right) &=\mathbb{E}_{\tau \sim \pi_{\beta}(\tau)}\left[\frac{\pi_{\theta}(\tau)}{\pi_{\beta}(\tau)} \sum_{k=0}^{H} \gamma^{t_{k+1}-t_0} r_k\right] \\
&=\mathbb{E}_{\tau \sim \pi_{\beta}(\tau)}\left[\sum_{k=0}^{H}\left(\prod_{k^{\prime}=0}^{H} \frac{\pi_{\theta}\left(a_{k^{\prime}} \mid s_{k^{\prime}}\right)}{\pi_{\beta}\left(a_{k^{\prime}} \mid s_{k^{\prime}}\right)}\right) \gamma^{t_{k+1}-t_0} r_k\right].
 \end{aligned}
 \end{align}

Since $\tau_k$ does not depend on $s_{k^{\prime}}$ and $\mathbf{a}_{k^{\prime}}$ for $t^{\prime} > t$, we can drop the importance weights from future time steps, resulting in the per-decision importance sampling estimator~\cite{precup2000eligibility} as given below.

\begin{equation}
 \begin{aligned}
J\left(\pi_{\theta}\right) &= \mathbb{E}_{\tau \sim \pi_{\beta}(\tau)}\left[\sum_{k=0}^{H}\left(\prod_{k^{\prime}=0}^{k} \frac{\pi_{\theta}\left(a_{k^{\prime}} \mid s_{k^{\prime}}\right)}{\pi_{\beta}\left(a_{k^{\prime}} \mid s_{k^{\prime}}\right)}\right) \gamma^{t_{k+1}-t_0} r_k\right] \\
&\approx \frac{1}{n} \sum_{i=1}^{n} \sum_{t=0}^{H} w_{k}^{i} \gamma^{t_{k+1}-t_0} r_{k}^{i},
 \end{aligned}
\end{equation}

where $w_{k}^{i}=\frac{1}{n} \prod_{k^{\prime}=0}^{k} \frac{\pi_{\theta}\left(a_{k^{\prime}}^{i} \mid s_{k^{\prime}}^{i}\right)}{\pi_{\beta}\left(a_{k^{\prime}}^{i} \mid s_{k^{\prime}}^{i}\right)}$. Such estimator can be utilized for policy evaluation. We can evaluate and pick the best policy with respect to their estimated returns.

For offline evaluation, we may suffer from the curse of dimensionality of the large state space and the curse of the long horizon in very large and highly mixed offline dataset. Therefore, tracing the policy outcomes along the time horizon to construct the weight $w_{k}^{i}$ is not applicable and could lead to very large variance~\cite{yuan2022notification} due to the accumulative product of $\prod_{k^{\prime}=0}^{k} \frac{\pi_{\theta}\left(\mathbf{a}_{k^{\prime}}^{i} \mid s_{k^{\prime}}^{i}\right)}{\pi_{\beta}\left(a_{k^{\prime}}^{i} \mid s_{k^{\prime}}^{i}\right)}$. As a trade-off, in this paper, we adopt the importance sampling with one-step approximation~\cite{chen2019top} to construct a biased but low variance estimator by setting $w_{k}^{i}= \frac{\pi_{\theta}\left(a_{k}^{i} \mid s_{k}^{i}\right)}{\pi_{\beta}\left(a_{k}^{i} \mid s_{k}^{i}\right)}$. This one-step approximation method is a fast and reliable method for offline evaluation for large-scale offline datasets. Comparison of the one-step approximation with other methods is provided in \cite{yuan2022notification}.

\section{System Architecture}
\label{sec:system}
In this section, we describe the system architecture for training and deploying RL policies. We carry out offline and online experiments in this system in Section~\ref{sec:experiments}. This system is fully deployed at LinkedIn and serves hundreds of millions of users.

\begin{figure}
    \includegraphics[width=0.5\textwidth]{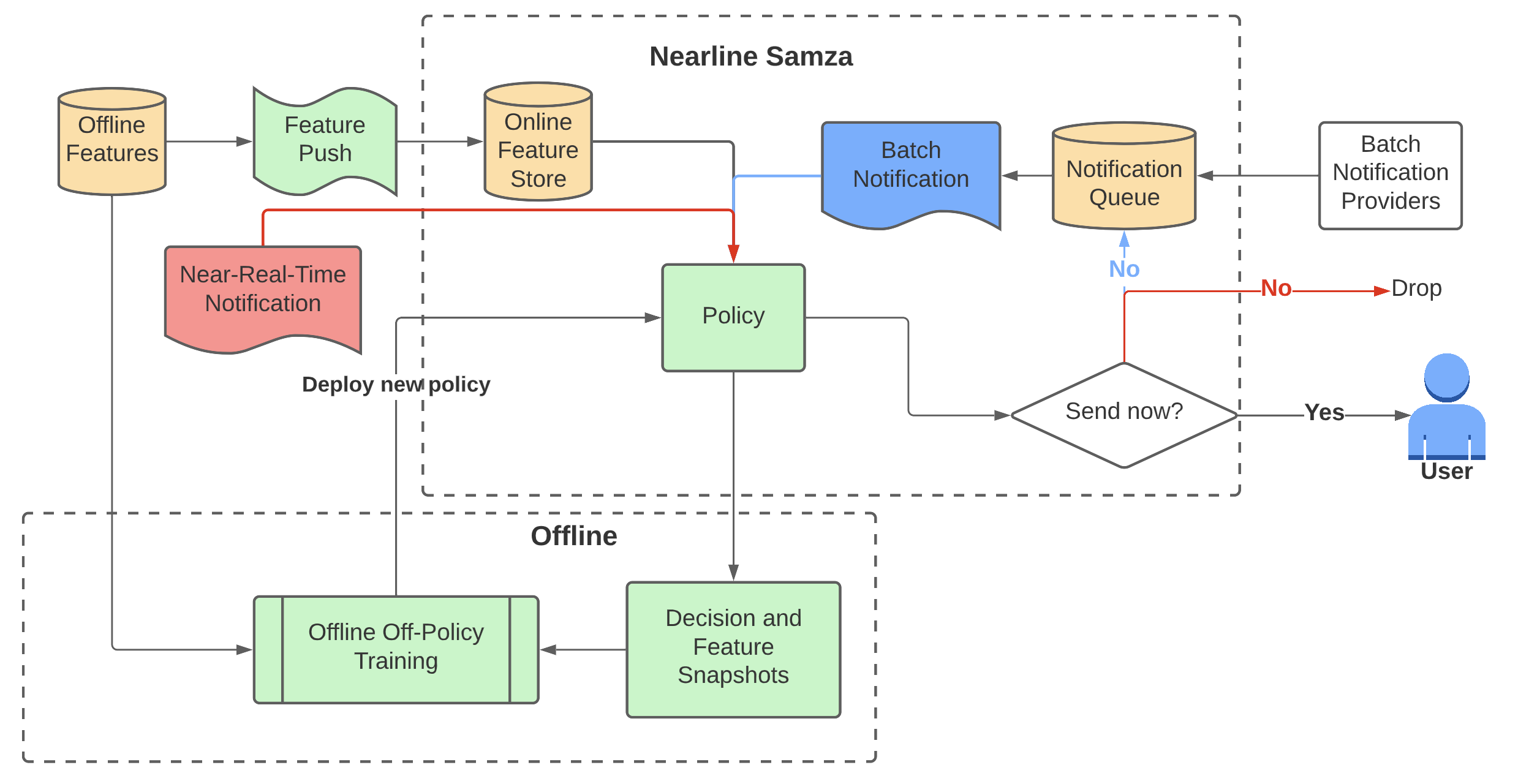}
     \caption{System architecture}
     \label{fig:system_archi}
\end{figure}

Figure \ref{fig:system_archi} shows the architecture of the notification decision system. In addition to the near-real-time notifications, this near-real-time system can also process batch notifications through a queuing system (one queue for each user), in which batch notification candidates for the user are queued. At fixed time intervals (a few hours), the top-quality notification in the queued notifications will form a near-real-time request to the policy just like a near-real-time notification. With the queuing system, we  spread out the request time for different users to avoid overloading the near-real-time system with a big batch. The policy is the decision-making module that yields SEND or NOT-SEND actions on each notification request. If the action is SEND, the evaluated notification will be delivered to the user. For near-real-time notifications, NOT-SEND action means dropping the notification. The batch notification may be put back into the queue if the decision is NOT-SEND since they are not time-sensitive and can be delivered later. To train the RL policy in this system,  we first collected one-week snapshot data from a small percentage of LinkedIn users using an epsilon-greedy behavior policy on top of a baseline policy previously deployed. The data was then joined with other log data to construct the tuples ($s_{t}$, $a_{t}$, $s_{t+1}$, $r_{t}$) for offline training. Important state features include the app badge count, the user’s last visit time, the number of notifications received over the past week and other user profile features. We then train the Conservative Double DQN models described in Section \ref{sec:dqn} in TensorFlow using a fully-connected 3-layer neural network. We usually train multiple polices based on different hyperparameters and reward preference weights. We then apply the offline policy evaluation to study offline or to choose candidates for online experiments. 


All RL policies are served in a nearline Samza service \cite{noghabi2017samza} using TensorFlow Serving for the TensorFlow model.  A policy takes online features, makes decisions, and then snapshots its decisions and features to a Hadoop Distributed File System (HDFS) \cite{shvachko2010hadoop}. Additional offline features can be used for offline training and pushed to Samza data stores for nearline serving.

\section{Experiments}
\label{sec:experiments}

In this section, we describe offline and online experiments that demonstrate the technologies discussed in previous sections. For offline RL, offline policy performance evaluation plays an important role as online testing is often limited by constraints on user experience. We first show that the offline policy evaluation method we introduced above provides sufficient consistency with online policy evaluation. We thus use offline evaluation for many studies below. We also show online results where available. Reward function \eqref{eq:linear_rewards_prediction} is used unless explicit called out otherwise.

 We are interested in user site engagement and notification engagement, which can be characterized by the following metrics. 

• \textbf{Sessions}: A session is a collection of full-page views made by a single user on the same device type. Two sessions are separated by 30 minutes of zero activity. This is a widely used metric on user engagement across social
networks.

• \textbf{Weekly Active Users (WAU)}: The number of users who have at least one session in a given week. We use WAU as a longer-term site engagement measure in comparison to sessions.

• \textbf{Notification Volume}: The total number of notifications served to users after removing potential duplicates. 

• \textbf{Notification CTR}: This metric measures the average click-through-rate of notifications sent to a user in a day. This is a metric capturing notification quality.

For online experiments, selected policies based on the offline evaluation results were deployed in the Samza service on a certain percentage of total users for an online A/B test compared with the baseline policy. To avoid disclosing sensitive information, we hide the label values in the figures.

\subsection{Validation of Offline Evaluation}
\label{sec:validateOfflineEvaluation}
\begin{figure}
    \includegraphics[width=0.5\textwidth]{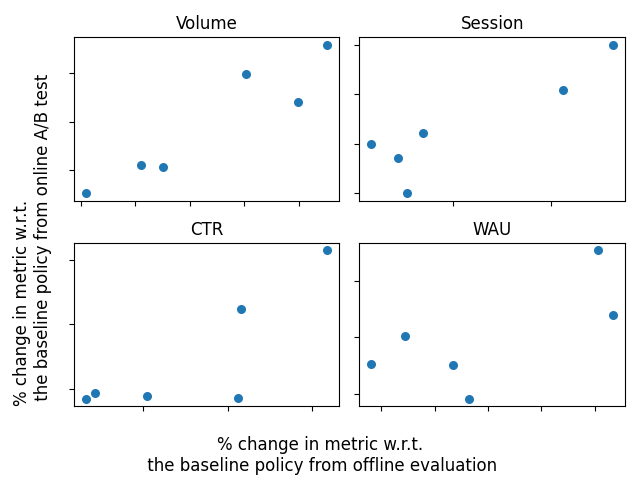}
    \caption{Validation of offline evaluation}{Pearson correlation $r^2=0.94, 0.90, 0.82, 0.68$ respectively for Volume, Session, CTR and WAU}
    \label{fig:online-offline}
\end{figure}

To validate offline evaluation as a reasonable method for policy evaluation, we compared the online and offline performance for 6 policies on the metrics notification volume, sessions, notification CTR and WAU. Figure \ref{fig:online-offline} shows the results. For the reasons discussed in section~\ref{sec:offlineEvaluation}, the offline evaluation with one-step approximation method carries biases that systematically overestimates the metrics compared to the online A/B test results. However, the two values are in general linearly correlated (see the Pearson correlation coefficients in the figure~\ref{fig:online-offline} caption), which indicates the hyperparameter tuning through offline evaluation could pick the right policy expected to perform well online. 

\subsection{Offline Experiments on CQL}
\label{sec:offlineExperiments}

\begin{figure}
    \begin{subfigure}[t]{.5\textwidth}
      \includegraphics[width=.46\linewidth]{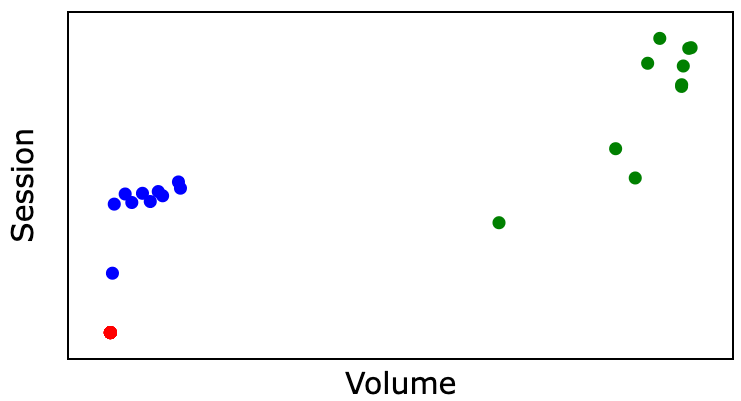}
      \includegraphics[width=.46\linewidth]{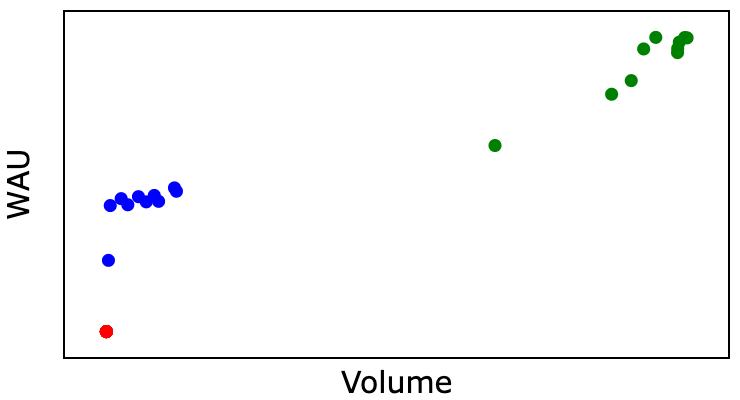}
     \end{subfigure}
     \begin{subfigure}[t]{.25\textwidth}
    
    \end{subfigure}
    \begin{subfigure}[t]{.5\textwidth}
    \includegraphics[width=.46\linewidth]{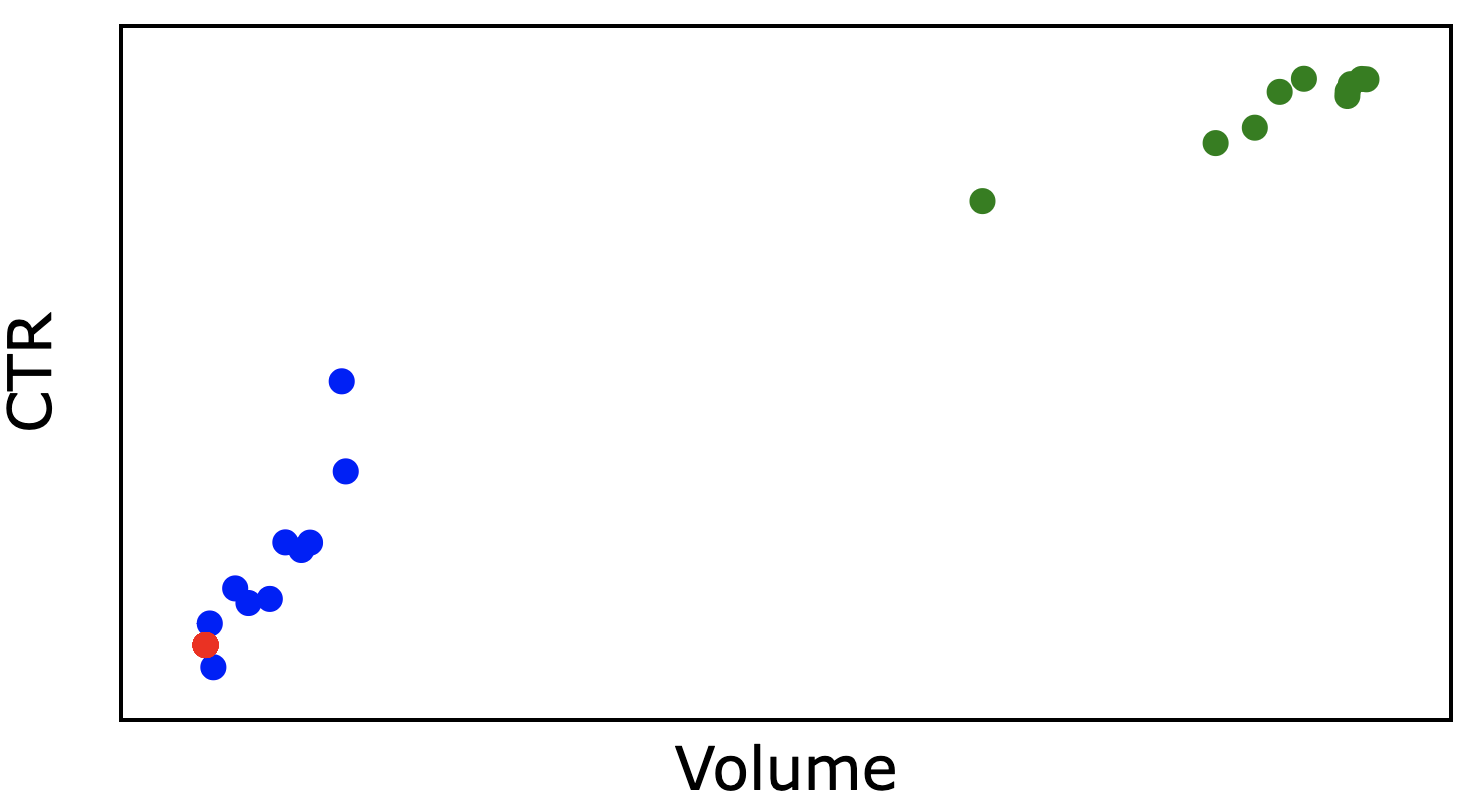}
    \end{subfigure}
     \caption{Performance of CQL}{Red dot: values calculated based on the baseline policy; blue dots: metric performance of DDQN models with CQL; green dots: metric performance from DDQN models without CQL.}

     \label{fig:cql}
\end{figure}

Next, we evaluate applying CQL on DDQN models. The goal is to obtain policies that can make optimal notification decisions such that the volume of notifications sent is controlled and key metrics such as Sessions, WAU and CTR are improved. We use predefined preference weights for the target rewards of sessions, expected CTR (CTR predicted from a supervised learning model) and volume, and trained two sets of double DQN models, one set with the CQL penalty term (test group) and the other set without the CQL penalty term (control group). Offline evaluation (Figure \ref{fig:cql}) shows that RL models with the CQL penalty term learned to increase the key metrics by more than 10\% while controlling volume (volume increased between 5\% to 25\%) for most of the learned policies. Moreover, the test group with CQL outperforms the control group across all metrics. As we can see from Figure \ref{fig:cql}, not only are the performance of different policies across multiple training runs more consistent, but also the resulting policies are more conservative with respect to the baseline policy (in terms of notification volume).

\subsection{Training with Predicted Sessions}
\label{sec:predicted_delta_visit}
\begin{figure}
  \centering
  \includegraphics[width=.7\linewidth]{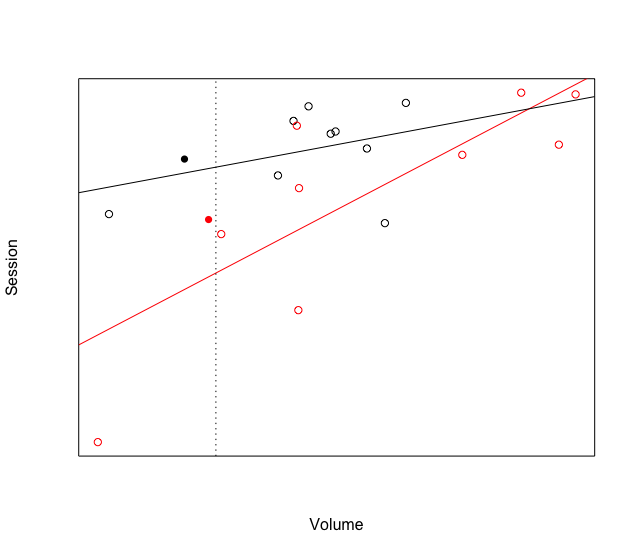}  
 \caption{Training with predicted rewards}{Black dots (solid and hollow): offline evaluation from training with reward function \eqref{eq:linear_predicted_delta_visit}; red dots (solid and hollow): offline evaluation from training with reward function \eqref{eq:linear_rewards_prediction}; Black and red dots in solid color: policies picked for online experiments; vertical line: the volume estimate of production notification system }
\label{fig:predicted_delta_session}
\end{figure}

In this experiment, we test training with the predicted sessions described in \cite{yuan2019state}. We denote its predicted value as $\hat{E}(m^{s}_k|a_k,s_k)$. We train RL with the following reward function

\begin{equation}
\label{eq:linear_predicted_delta_visit}
    r_k =  \omega^{s}_k \hat{E}(m^{s}_k|a_k,s_k)  + \omega^{c}_k \hat{E}(m^{c}_k|a_k,s_k)  + \omega^{v}_k m^{v}_k,
\end{equation}
and compare to training with reward function provided in \eqref{eq:linear_rewards_prediction}. The training with predicted sessions outperforms its counterpart on the session evaluation in most of the comparable region on notification volume. To confirm this offline finding, we choose two polices (denoted as solid dots in Figure \ref{fig:predicted_delta_session}) for the online experiment in Section~\ref{sec:online_predicted_delta_pvisit}.

\subsection{Influence Reward Trade-offs through Preference Function Setting}
\label{sec:prefVector}
\begin{figure}
    \includegraphics[width=.8\linewidth]{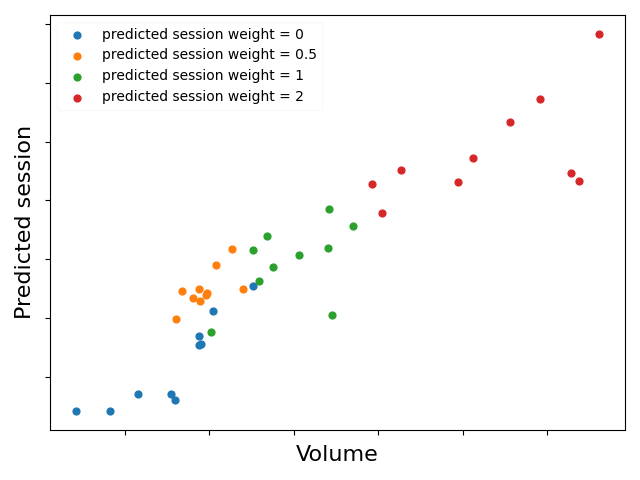}
     \caption{Influence reward trade-off through preference function setting}
     \label{fig:pref_func}
\end{figure}

When optimizing for multiple rewards, there might be a need to influence the learning process so that the learned policy emphasizes on one reward more than another. We show that this can be done intuitively through adjusting the weights for individual rewards in the preference function. For this, we tested four different reward weights $\omega^{s}_k$ for the predicted sessions objective in a MORL setting in \eqref{eq:linear_predicted_delta_visit} while keeping the reward weights on the other two objectives and all other hyperparameters the same. To account for the inherent randomness in the learning process, we repeated the training for each preference function 10 times. In Figure \ref{fig:pref_func}, we show the offline evaluation results for the three metrics for the trained policies on the validation dataset. When $\omega^{s}_k$ increases, the policies explore regions of higher sending volumes to gain higher predicted session rewards, trading off with lower volume reward.

\subsection{Online Experiments in the Near-real-time Notification System}
\label{sec:onlineExperiments}
We deploy proposed RL models in the near-real-time notification system in LinkedIn. We use the best-performing MOO model, optimizing the same set of objectives as described in \cite{amuralidharan2022}, as the baseline. The baseline uses supervised learning models, which predict the impact of the send decision on the objectives of interest, sessions ($ \hat{E}(m^{s}_k|a_k,s_k)$) and clicks ($\hat{E}(m^{c}_k|a_k,s_k)$). This approach solves a MOO with a decision function of choosing SEND if $ \hat{E}(m^{s}_k|a_k,s_k)+\beta  \hat{E}(m^{c}_k|a_k,s_k)> \lambda(.)$, where $\beta$ and $\lambda(.)$ are derived from the duality of the MOO. The term $\lambda$, which represents a send threshold function in the baseline model is used to control volume and the trade-offs between the short-term and long-term rewards. The baseline model considered parameterized threshold functions as described in \cite{amuralidharan2022}, which provided online improvements on top of a simple scalar threshold. The parameters $\beta$ and the parameters of $\lambda(.)$ were obtained using online bayesian hyperparameter tuning. Under the framework of A/B tests, we compare both Double DQN with and without conservative q-learning loss to the baseline. We summarize the key metrics in Table \ref{tab:abtest_cql}. All reported numbers are statistically significant with p-value $\le 0.05 $. Table \ref{tab:abtest_cql} shows that Double DQN without the conservative q-learning loss does not achieve as good sessions-volume and CTR-volume trade-offs in comparison to the baseline. The conservative q-learning loss makes a significant difference to the offline learning efficiency and improves online metrics over the baseline. While decreasing volume, thereby sending fewer notifications, the Double DQN model with CQL improves both Sessions and WAU, which measures the longer-term site engagement. The reported gains in Sessions and WAU are percentage changes to site-wide overall numbers in comparison to the baseline, including organic sessions and WAU that are not driven by notifications. These two metrics are usually harder to move without increasing Volume than CTR metric, and hence $+0.24\%$ sessions and $+0.18\%$ WAU are considered significant business impact.  We have ramped the DDQN + CQL model to all users based on this result.

\begin{table}[!h]
\caption{Online A/B test results for DDQN with and without CQL}
\centering
  \begin{tabular}{ | l | c | c| }
    \hline
    Metric & DDQN vs. Baseline & DDQN + CQL vs. Baseline\\
    \hline
    \hline
   Sessions & not stat sig & + 0.24\% \\
    \hline
    WAU & -0.69\% & + 0.18\% \\
     \hline
    Volume  & +7.72\%& -1.73\% \\
    \hline
    CTR & -7.79\% & +2.26\% \\

    \hline
  \end{tabular}
\label{tab:abtest_cql}
\end{table}

\subsection{Online Experiments for the Predicted Sessions} \label{sec:online_predicted_delta_pvisit}
Table~\ref{tab:abtest_session} shows the online A/B test results comparing two policies trained using the observed and the predicted session rewards respectively, corresponding to the solid dots in Figure~\ref{fig:predicted_delta_session}. The results are directionally consistent with offline evaluation. The policy trained with predicted sessions achieves more sessions with smaller notification volume, with an increase in CTR as well.

\begin{table}[!h]
\caption{Online A/B results using predicted sessions}
\centering
  \begin{tabular}{ | l | c | c| }
    \hline
    Metric & Predicted sessions vs. Sessions\\
    \hline
    \hline
   Sessions & +0.35\%  \\
    \hline
    WAU & neutral\%  \\
     \hline
    Volume  & -1.44\%\\
    \hline
     CTR & +2.37\% \\
  
    \hline
  \end{tabular}
\label{tab:abtest_session}
\end{table}

\section{Discussion}
In this paper, we propose an offline RL approach for large-scale near-real-time notification systems. We argue that RL is a principled approach to optimize for a sequence of well-coordinated notification decisions. We demonstrate that RL can achieve multi-objective trade-offs through reward engineering. We suggest a potential improvement in learning by leveraging reward prediction models, although its theoretical justifications remain to be studied. 

Compared with online RL, offline RL is better suited for such real-world applications, but also presents unique challenges to offline learning and offline evaluation. We demonstrate that popular algorithms such as DDQN may not work as well in the offline setting. CQL, which is designed for offline learning is our choice. We demonstrate in the online A/B test that CQL performs better in a multi-objective set-up than the most popular MOO approach based on supervised learning. 

For future work, we plan to investigate how to efficiently do offline evaluation with approximators that have minimal bias while keeping the variance low. We also want to explore more sophisticated methods, such as the non-linear scalarization functions to improve the MORL framework and obtain a better spread amongst the set of Pareto optimal solutions~\cite{luc2008pareto}.



\begin{acks}
This work was supported by LinkedIn Corporation. We thank our colleagues: Gao Yan, Ankan Saha and Preetam Nandy for reviewing the manuscript and their insightful suggestions on this work. We also thank the anonymous reviewers for their valuable comments and suggestions.
\end{acks}

\bibliographystyle{ACM-Reference-Format}
\bibliography{morl_kdd}

\appendix

\end{document}